\documentclass{article}
\usepackage{spconf,amsmath,graphicx}

\title{EAA-Net: Rethinking the Autoencoder Architecture with Intra-class Features for Medical Image Segmentation}
\name{Shiqiang Ma\textsuperscript{\rm 1, \rm \&}, Xuejian Li\textsuperscript{\rm 1, \rm \&}, Jijun Tang\textsuperscript{\rm 2, \rm *} , Fei Guo\textsuperscript{\rm 3, \rm *} }
\address{\textsuperscript{\rm 1}School of Computer Science and Technology, College of Intelligence and Computing,\\ Tianjin University, China\\
\textsuperscript{\rm 2}Shenzhen Institute of Advanced Technology, Chinese Academy of Sciences, China \\
\textsuperscript{\rm 3}School of Computer Science and Engineering, Central South University, China \\
\textsuperscript{\rm *}Corresponding Authors: Jijun Tang (tangjijun@tju.edu.cn),  Fei Guo (guofei@csu.edu.cn) \\
\textsuperscript{\rm \&}These authors contributed equally to this work and should be considered co-first authors.}

\begin{document}
\topmargin=0mm
%
\maketitle
\begin{abstract}
Automatic image segmentation technology is critical to the visual analysis. The autoencoder architecture has satisfying performance in various image segmentation tasks. However, autoencoders based on convolutional neural networks (CNN) seem to encounter a bottleneck in improving the accuracy of semantic segmentation. Increasing the inter-class distance between foreground and background is an inherent characteristic of the segmentation network. However, segmentation networks pay too much attention to the main visual difference between foreground and background, and ignores the detailed edge information, which leads to a reduction in the accuracy of edge segmentation. In this paper, we propose a light-weight end-to-end segmentation framework based on multi-task learning, termed Edge Attention autoencoder Network (EAA-Net), to improve edge segmentation ability. Our approach not only utilizes the segmentation network to obtain inter-class features, but also applies the reconstruction network to extract intra-class features among the foregrounds. We further design a intra-class and inter-class features fusion module -- $I^{2}$ fusion module. The $I^{2}$ fusion module is used to merge intra-class and inter-class features, and use a soft attention mechanism to remove invalid background information. Experimental results show that our method performs well in medical image segmentation tasks. EAA-Net is easy to implement and has small calculation cost.
\end{abstract}
\begin{keywords}
Medical image segmentation, multi-task learning, COVID-19-20 detection, intra-class features
\end{keywords}
\section{Introduction}
\label{sec:intro}

Image segmentation is a very important part of computer vision. It separates the foreground from the background. This is an important work for image understanding and analysis. Traditional image segmentation methods \cite{1} required manual intervention when extracting features, which resulted in algorithm performance being controlled by human subjective factors. Image segmentation based on deep learning did not require human intervention, which adaptively extracted feature vector, and approximately mapped to the objective function through a deeper level and a nonlinear activation function. Fully convolutional networks (FCNs) \cite{3} used a convolutional layer instead of fully connected layer, which could perform pixel-level classification, FCNs solved the problem that fully connected layers lose spatial context information, making the deep learning models more suitable for image segmentation tasks. The autoencoder architectures \cite{77} based on FCNs achieved encouraging performance on image segmentation tasks and have been widely used in the field of medical image segmentation \cite{6}\cite{12}, target detection \cite{9} and video object segmentation \cite{56}. Autoencoders obtained the high-level representations of the input data. In the latter part of the neural network, these high-level information were used for different tasks. The down-sampling method in autoencoder expanded the receptive field while reducing the number of network parameters, but down-sampling caused the loss of edge information. Besides, the segmentation network paid more attention to the mainly different area between foreground and background, which caused the model to ignore the edge details.

\begin{figure*}[htbp]
	\centering
	\includegraphics[scale=1.75]{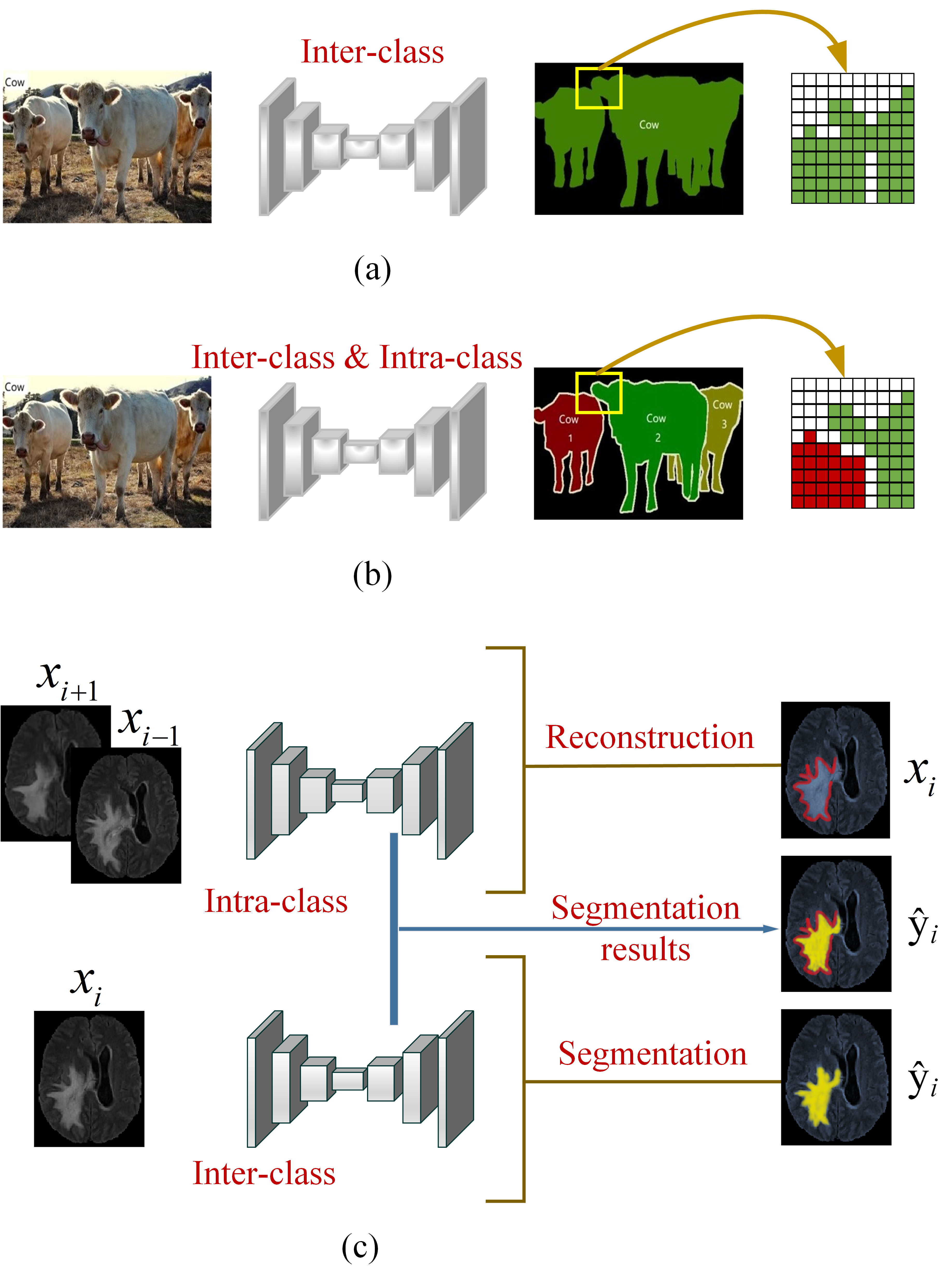}
	\caption{Inspired by the difference between semantic segmentation and instance segmentation, our method combines the intra-class and inter-class features to obtain accurate edge segmentation results. $x_{i}$ represents the image of the $i^{th}$ frame, $\hat{y}_{i}$ represents the label of the $i^{th}$ frame. Red indicates the area where the edge changes between different frames. (a) Semantic segmentation methods. (b) Instance segmentation methods. (c) EAA-Net.}
	\label{fig1}
\end{figure*}

To solve the problem, HRNet \cite{14} connected high-resolution and low-resolution features in parallel. Although this method has been proven to be effective, the increasing of parameters makes the network more difficult to be trained. U-Net \cite{4}, a U-shaped autoencoder structure, was applied to the task of image segmentation. It used skip connections to combine the features of encoder path and decoder path, so that the high-level layers obtained multi-scale features. Chen et al. proposed the method of Atrous Convolution \cite{16}, which enlarged the receptive fields without changing the resolution of image, and obtained multi-scale information. However, Atrous Convolution could lose local information when the atrous rate is large, and the input signal is sampled at a long-distance lack correlation. Valanarasu et al. presented a complete architecture named KiU-Net \cite{17}, which added an over-complete segmentation architecture Ki-Net in parallel to the under-complete convolutional autoencoder U-Net to capture edge information. Ki-Net required to upsample the input multiple times, which undoubtedly increased the computational cost and training difficulty. In addition, multiple convolution operations using the same loss as the U-Net branch make Ki-Net inevitably learn high-level semantic features rather than edge details. Attention U-Net \cite{18} added an attention module to the skip connections of U-Net. This attention module combined the corresponding feature maps in the encoder path and decoder path, and weighted the fusion results to control the importance of different spatial location features. Attention U-Net improved the low-to-high feature transfer capability of the U-Net architecture, but it did not solve the intentional ignorance of edge details by the segmentation network. To force the network to learn border pixels, U-Net proposed a weighted cross entropy with a pixel-wise loss weight. Similarly, many loss functions have been proposed to help segment the boundary between the foreground and the background \cite{21}. However, paying attention to the boundary between foreground and background cannot completely solve the problem of blurred edge, and it also reduces the ability of network to extract inter-class features.

The key to improving the edge segmentation performance of the model based on the autoencoder architecture in the 3D image segmentation task is to learn both the intra-class and inter-class features at the same time. BriNet \cite{55} used Information Exchange Modules, multi-path fine-grained strategy and online refinement strategy to solve the intra-class gap caused by insufficient information interaction between images, and the inter-class gap caused by the lack of overlap between object categories in the training and prediction phases. Simply learning boundary features cannot solve the problem of blurred edges. Learning the intra-class features makes the model focus on the edge details, which is the best choice to intelligently solve the edge blur.

Inspired by the difference between semantic segmentation and instance segmentation, we propose a novel multi-task architecture EAA-Net that combines intra-class features with inter-class features, which is illustrated in Fig. \ref{fig1}. Unlike the instance segmentation method, our approach learns the difference between a target in different frames. The difference of same target in neighboring slices reveals its edge change law, Fig. \ref{fig6} shows the difference in the foreground regions of adjacent images. We use an reconstruction network to reconstruct the image that needs to be segmented. As shown as Fig. \ref{fig1} (c), we use the image of the $i+1^{th}$ frame and the $i-1^{th}$ frame to reconstruct the $i^{th}$ frame. In this way, The reconstruction network learn the law of the edge change of the target between neighboring slices. We think it is a special intra-class feature between different frames. Through learning the law of edge change, the reconstruction network can focus on the edge details of the target. The segmentation network is used to learn the inter-class features between foreground and background. In addition, we design an $I^{2}$ fusion module that integrates the inter-class and intra-class features. The $I^{2}$ fusion module acts as a bridge to connect two completely independent visual representation features. It can avoid the influence of the background on the segmentation task. The main contributions of this paper are as follows:

\begin{figure}[htbp]
	\centering
	\includegraphics[scale=0.5]{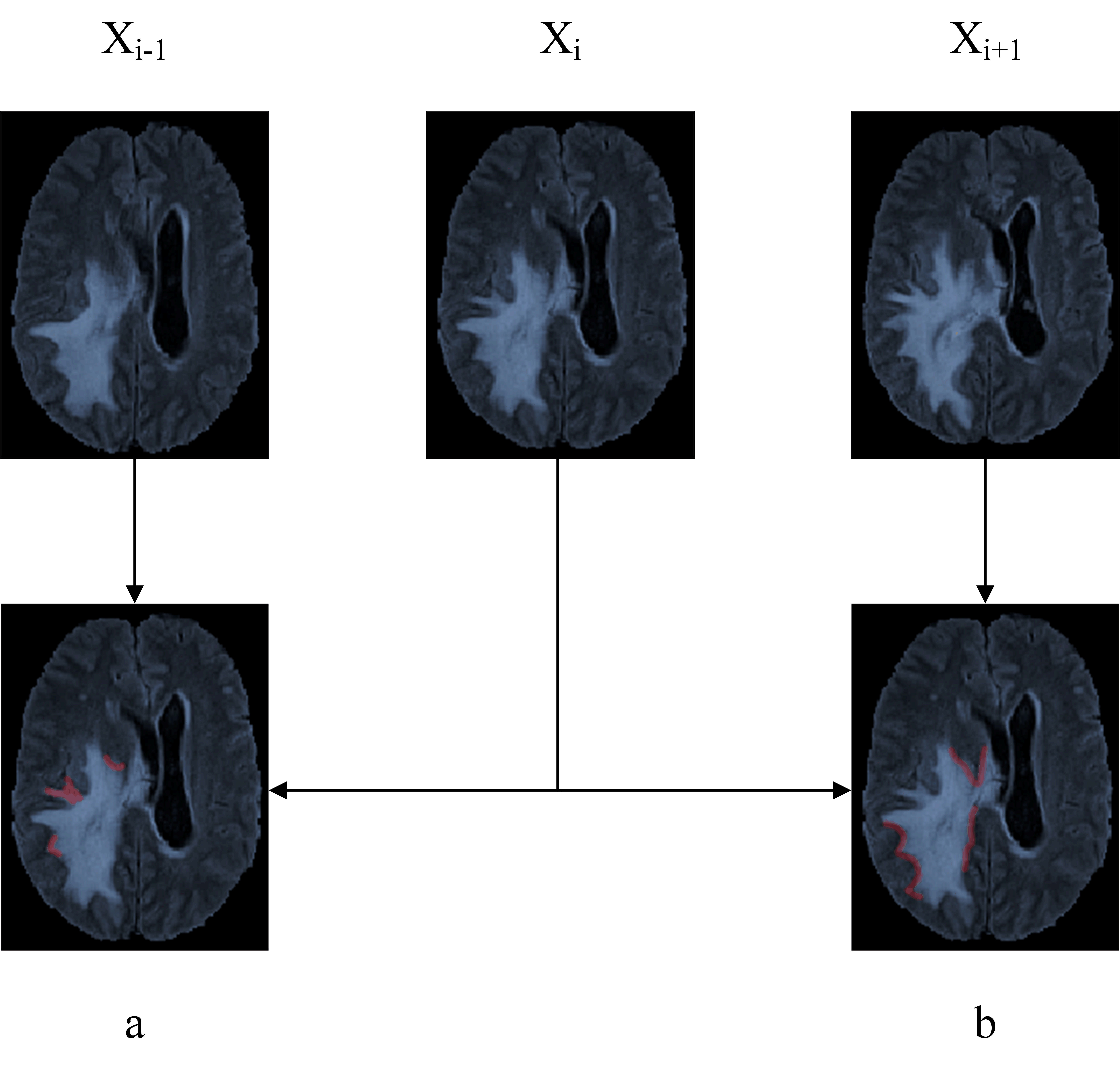}
	\caption{Differences in foreground regions of adjacent images. (a) The red areas illustrate the difference between the foreground regions of $x_{i-1}$ and $x_{i}$. (b) The red areas show the difference between the foreground regions of $x_{i+1}$ and $x_{i}$.}
	\label{fig6}
\end{figure}

\begin{itemize}
\item We propose EAA-Net, a novel edge attention autoencoder architecture. EAA-Net helps autoencoder-based segmentation models overcome their inherent drawbacks (downsampling operations lead to loss of edge details).
\item We design a multi-feature fusion method named $I^{2}$ fusion module, which is used to fuse the features of inter-class and intra-class. $I^{2}$ achieves a satisfying performance, and can be applied to a variety of feature fusion scenarios.
\item We have extensively verified the effectiveness of EAA-Net for medical image segmentation tasks. EAA-Net can use any high-performance autoencoder architecture as its backbone and improve the quality of segmentation edges.
\end{itemize}

\section{Related Work}

Deep learning methods achieved impressive results benefited from a huge labeled data. However, there are not enough trainable labeled data for many tasks in the real world, like medical imaging \cite{24} and medical image segmentation \cite{83}. A small amount of labeled data inevitably lead to the overfitting problem and reduced the generalization ability of the trained model. Although transfer learning strategy dealt with the novel problem, the performance of transfer learning expected to decline, when there was a big difference between the distribution of source data and target data \cite{25}. In this case, multi-task learning can help alleviate this data sparsity problem by using useful information from other related learning tasks. The goal of multi-task learning is to use the useful information contained in multiple learning tasks to help the model learn richer features. According to the task, multi-task learning can be divided into multi-task supervised learning, multi-task unsupervised learning \cite{29, 30, 31}, multi-task semi-supervised learning \cite{34, 35}, etc. 

Generally, the labeled data and unlabeled data were simultaneously used for training by semi-supervised learning. Therefore, semi-supervised learning combines the advantages of supervised learning and unsupervised learning. Mean Teacher \cite{41} used consistency loss to make the distribution of the network output of the same original image that have been randomly enhanced as similar as possible. Though extracting a variety of visual features, the network learned to understand the content expressed by image. It made training with a small number of labeled datasets more robust. In order to cope with more visual interference, the data enhancement method proposed by UDA \cite{43} made full use of unlabeled data in the case of a small number of samples, so that the network performance after training reached the high accuracy of training with a fully labeled dataset. Some methods introduced virtual adversarial training (VAT) \cite{44} or Pseudo-Labels \cite{45} in self-supervised learning to semi-supervised learning, and used the loss function based on unsupervised technology to train the unsupervised part separately.

Through performing specific tasks, unlabeled data played an important role in training process. However, self-supervised learning and semi-supervised learning strategies required the generation of pseudo-labels by manual labeling. Usually, obtaining valuable pseudo-labels required more labor cost and time cost, which undoubtedly increased the difficulty of training the model. In addition, human-designed pseudo-labels were subjective and uncertain, making it difficult to be widely applied to multiple tasks. Some simple and effective multi-task learning methods were proposed to improve the detrimental effects of using only a small number of samples on medical image analysis tasks. Andriy Myronenko \cite{78} designed a multi-task learning method for the brain tumor segmentation task, which used the reconstruction task as the regularization term of the model to prevent the model from overfitting due to few training samples. Weninger et al. \cite{79} applied a multi-task learning strategy to the glioma segmentation task, and they used a classification task and a reconstruction task to assist the glioma segmentation task. By jointly training the above three tasks, the encoder part extracted richer features. In this paper, we proposed a lightweight multi-task learning method, which did not use complex multi-task learning strategies, but utilized the intra-class features between several adjacent images with spatial correlation to quickly and sharply capture edge details information, and assisted the model to accurately segment the region of interest.

\section{Methods}

\begin{figure*}[htbp]
	\centering
	\includegraphics[scale=0.95]{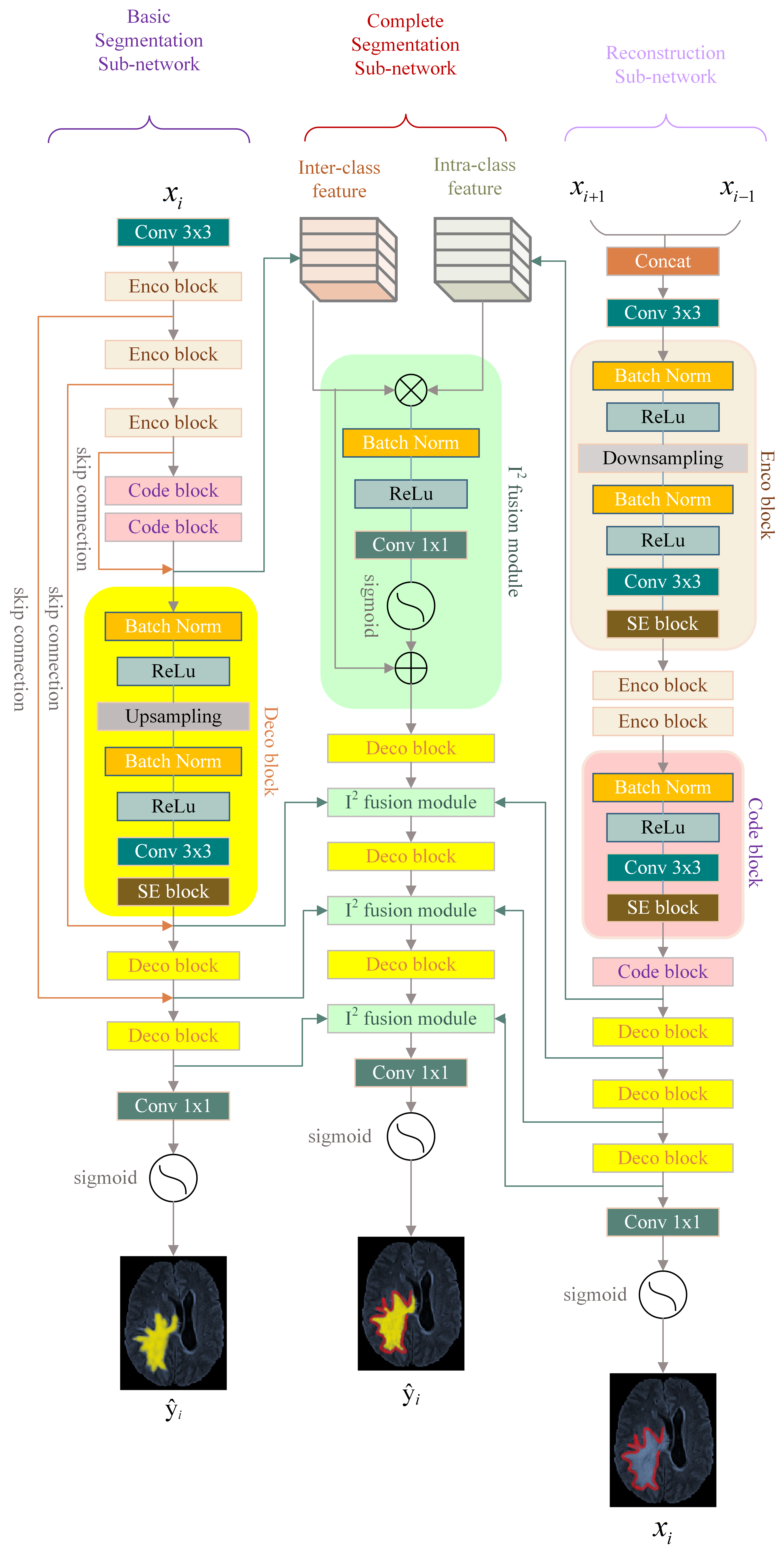}
	\caption{The overview of EAA-Net. EAA-Net is composed of Basic Segmentation Network, Reconstruction Network and Complete Segmentation Network. Here, both the basic segmentation network and the reconstruction network are autoencoder architectures. We design Enco block and Deco block in the encoding part and the decoding part respectively. The Complete Segmentation Network includes $I^{2}$ fusion module and Deco block. The $I^{2}$ fusion module integrates the features of inter-class and intra-class to improve the quality of segmentation edges. $x_{i}$ represents the image of the $i^{th}$ frame, $\hat{y}_{i}$ represents the label of the $i^{th}$ frame.}
	\label{fig2}
\end{figure*}

In this section, we describe our approach in details. The architecture of EAA-Net can be seen in Fig. \ref{fig2}. EAA-Net consists of three branches, namely basic segmentation sub-network, reconstruction sub-network, and complete segmentation sub-network. Artificial intelligence is a technology in which machine mimics human "cognitive" functions. The operation mode of EAA-Net is similar to human brain \cite{75}. The basic segmentation sub-network is similar to the left brain for logical calculation. The reconstruction sub-network is similar to the right brain for creation, and makes up for the deficiencies of left brain. They are both independent and complementary. Complete segmentation sub-network with $I^{2}$ fusion module is similar to the corpus callosum for connecting two cerebral hemispheres, and merge their ‘ideas’. We will introduce our framework in more detail below.

\subsection{Basic Segmentation Sub-network}

EAA-Net can directly utilizes the classic autoencoder architectures as the backbone of the basic segmentation sub-network, such as U-Net and SegNet \cite{56}. Inspired by light-weight semantic segmentation architectures \cite{58}, we design a light-weight basic segmentation sub-network with a small parameters. 

The light-weight basic segmentation sub-network is similar to U-Net. We retain the contraction path (encoder), expansion path (decoder) and skip connections. Basic segmentation network consists of Enco block, Code block and Deco block. In addition, we add the channel attention mechanism--Squeeze-and-Excitation module (SE module) \cite{59}. The SE module weights the contribution of each channel to weaken the negative influence of some channels for improving the efficiency of the model. We also use batch normalization before the convolution layer and activation function. Let $x_{i}^{BE}$ be the output of $i^{th}$ encoder layer of basic segmentation network:
\begin{equation}
x_{i}^{BE} = Conv * \sigma_{1}(BN_{\gamma, \beta}(x_{i-1}^{BE})), \\ i\in\{i, \cdots, l\}, \label{eq:n1}
\end{equation}
where $BN_{\gamma, \beta}$ is batch normalization with the mean $\gamma$ and variance $\beta$, which normalizes the $x_{(i-1)}^{BE}$ to a gradient interval to avoid gradient vanishing or exploding. $\sigma_{1}$ represents activation function, which can increase the non-linearity and increase the robustness. $Conv$ denotes a 2D convolution with kernel size $3\times3$.

The input of the basic segmentation sub-network is a single frame of target image that needs to be segmented. Since the input of the model is 2D image with the batch size of $32$, batch normalization is more suitable as a normalization method for the model. We use deconvolution with a stride $2$ and a convolution with kernel size of $3\times3$ to perform up-sampling. Unlike the encoder path, the decoder path includes up-sampling and skip connection. Let $x_{i}^{BD}$ be the output of $i^{th}$ decoder layer of basic segmentation sub-network:
\begin{equation}
    \begin{split}
        x_{i}^{BD} = Conv * \sigma_{1}(BN_{\gamma, \beta} [TC(x_{i-1}^{BD}),  x_{i-1}^{BE}]), \\
        i \in \{i, \cdots, l-1\},
        \label{eq:n2}
    \end{split}
\end{equation}
where $x_{i-1}^{BE}$ is low-level feature maps from the encoder path with the same resolution as $x_{i}^{BD}$. Experiments show that the performance of the decoder using transposed convolution $TC$ with strides $2$ is better than using linear interpolation. The basic segmentation sub-network is designed to extract inter-class features, which is used as part of the input of the complete segmentation sub-network.

Normally, the segmentation network pays more attention to the main difference between the foreground and the background area, thereby ignoring the edge information. In order to make the network learn more about edge features, we design a reconstruction network to reconstruct the image that needs to be segmented.

\subsection{Reconstruction Sub-network}

Multi-task learning \cite{76} is an area that has received much attention in pattern recognition and machine learning. Multi-task learning methods can significantly improve the efficiency of data utilization and effectively alleviate the overfitting problem of the model. Imaging techniques using MR and CT can acquire a large amount of three-dimensional image data with spatial correlation information, and the biggest difference between two adjacent images is the edge detail information. In order to assist the segmentation model to obtain richer edge detail information to address the loss of detail information caused by downsampling operations, we design a novel and elegant image reconstruction task. Specifically, we use $x_{i-1}$ and $x_{i+1}$ as input to reconstruct $x_{i}$. $x_{i-1}$ and $x_{i+1}$ adjacent to $x_{i}$ contain a large number of similar features, and their differences are mainly concentrated in edge detail information. The purpose of this reconstruction task is to make the model pay attention to the intra-class features between adjacent images, that is, the region of interest edge details. Therefore, the multi-task learning strategy does not require artificial construction of pseudo-labels, which can save a lot of labor cost and time cost. We use the method to extract more edge details of the interested region.

Edge details and high-level semantics are important factors in semantic segmentation tasks. The reconstruction sub-network is responsible for edge details. The branch takes two images $x_{i-1}$ and $x_{i+1}$ adjacent to the target image $x_{i}$ as input, which is similar to image inpainting \cite{65}. The label of the branch is the target image $x_{i}$. The difference of ROI in adjacent images is mainly concentrated on the edge of the foreground and background information. Using two adjacent images as input can provide rich edge information and reduce the interference of noise on the edge information of a single image.

The structure of the reconstruction sub-network is similar to that of the basic segmentation sub-network, which facilitates the integration of the features of two branches at the same resolution. We do not add skip connections in the branch to avoid the influence of low-level features of non-target images on the prediction results. The difference of intra-class features is smaller than inter-class, so we reduce the channel capacity of this branch. The experimental results show that the lightweight reconstruction network can effectively improve the segmentation performance and edge quality. This may be related to the different contributions of intra-class and inter-class features to segmentation task.

\subsection{Complete Segmentation Sub-network}

The complete segmentation sub-network is responsible for fusing features of inter-class and intra-class, and the low-resolution feature map is restored to the original size through deconvolution. The branch includes 2D convolution, batch normalization, ReLU, and $I^{2}$ fusion module. The output $x_{i}^{C}$ of $i^{th}$ feature of the complete segmentation sub-network can be expressed as:
\begin{equation}
    \begin{split}
        x_{i}^{C} = & Conv * \sigma_{1}(BN_{\gamma, \beta} (TC(x_{i-1}^{C}) \oplus \\ 
        & I^{2}(x_{i}^{BD}, x_{i}^{RD}))), 
        i\in\{i, \cdots, l\},
        \label{eq:n3}
    \end{split}
\end{equation}
where $I^{2}(x_{i}^{BD},x_{i}^{RD})$ is an edge attention module that combines the output $x_{i}^{BD}$ of the $i^{th}$ layer of the basic segmentation sub-network and the output $x_{i}^{RD}$ of the $i^{th}$ layer of the reconstruction sub-network. We use the element-wise addition method $\oplus$ to fuse the output $x_{i-1}^{C}$ of the previous layer of the complete segmentation sub-network after deconvolution and the output of the $I^{2}$ fusion module. The initial input $x_{0}^{C}$ of the complete segmentation sub-network is defined as:
\begin{equation}
x_{0}^{C} = I^{2}(x_{l}^{BE}, x_{l}^{RE}),
\label{eq:n4}
\end{equation}
where $x_{l}^{BE}$ and $x_{l}^{RE}$ are the output of last encoder layer of the basic segmentation sub-network and the reconstructed sub-network respectively. The details of the $I^{2}$ fusion module are illustrated as follows.

\subsection{Intra-class and Inter-class Fusion Module}

The features of the reconstruction sub-network and the basic segmentation sub-network are different in level of visual representation. In addition, the reconstruction sub-network will inevitably be affected by the background when extracting the intra-class features in the foreground. Therefore, we cannot simply merge these features. We design $I^{2}$ fusion module to fuse the features of intra-class and inter-class, and curb the influence of the backgrounds on the segmentation task.

$I^{2}$ fusion module includes element-wise multiplication, batch normalization, ReLU, $1\times1$ convolution, sigmoid, element-wise addition, which can be seen in Fig. \ref{fig2}. The inter-class features between the foreground and background extracted by the basic segmentation sub-network and the intra-class features of the target extracted by the reconstruction sub-network are used as the input of the $I^{2}$ fusion module. The output $I^{2}(x_{i}^{BD},x_{i}^{RD})$ of the $I^{2}$ fusion module is the edge-weighted feature maps which fuses the segmentation features and the reconstruction features:
\begin{equation}
x_{i}^{CT} = W_{i}^{T} \sigma_{1}(BN_{\gamma, \beta}(x_{i}^{BD}\odot x_{i}^{RD})), 
i\in\{i, \cdots, l\},
\label{eq:n5}
\end{equation}
\begin{equation}
    \begin{split}
        I^{2}(x_{i}^{BD}, x_{i}^{RD}) &= x_{i}^{BD} \oplus \sigma_{2}(x_{i}^{CT}) \\
        &= x_{i}^{BD} \oplus \sigma_{2}W_{i}^{T} \sigma_{1}(BN_{\gamma, \beta} \\
        & (x_{i}^{BD}\odot x_{i}^{RD})), 
        i\in\{i, \cdots, l\},
        \label{eq:n6}
    \end{split}
\end{equation}
where element-wise multiplication $\odot$ uses segmentation features $x_{i}^{BD}$ as a mask to reduce the effects of the background in the reconstruction feature $x_{i}^{RD}$ on the output. $W_{i}^{T}$ is a $1\times1$ convolution with the same size as the input, which is used to weight the new features that combine the features of intra-class and inter-class. The weighted features $x_{i}^{CT}$ are mapped to $(0,1)$ by the activation function sigmoid $\sigma_{2} = \frac{1}{1 + exp(x_{i}^{C})}$, and then merge with the feature $x_{i}^{BD}$ of the basic segmentation network through element-wise addition.

The $I^{2}$ fusion module is a soft attention method \cite{59} with a small parameters. It can be applied to the field of semantic segmentation with temporal or spatial contextual information. It is worth noting that the channel capacity of the reconstruction sub-network is smaller than the basic segmentation sub-network. Therefore, it is necessary to reduce the number of channels in the segmented feature maps or increase the number of channels in the reconstruction feature maps before introducing two features into the $I^{2}$ fusion module.

\subsection{Loss Function}

Our model includes three branches. We design a total loss function to make three independent branches collaborative. The total loss function is defined as follows:
\begin{equation}
Loss_{total} = Loss_{A} + Loss_{S} + Loss_{B} + Loss_{C},
\label{eq:n7}
\end{equation}
\begin{equation}
Loss_{A} = \frac{1}{n} \sum_{i=1}^{n} \vert y_{i} - \hat{y_{i}} \vert,
\label{eq:n8}
\end{equation}
\begin{equation}
Loss_{S} = \frac{1}{n} \sum_{i=1}^{n} (y_{i} - \hat{y_{i}})^{2},
\label{eq:n9}
\end{equation}
where $y_{i}$ and $\hat{y}_{i}$ are the ground-truth and prediction respectively. $Loss_{A}$ and $Loss_{S}$ are the loss functions of the reconstruction sub-network, they refer to MAE and MSE respectively, which are used to calculate the similarity between the prediction of the reconstruction sub-network and the real target image. The calculation of MSE is simple, and the penalties for abnormal points that deviate greatly are greater. However, MSE cannot effectively deal with abnormal points with small deviations. MAE has better robustness to deal with abnormal points with small deviations. Therefore, we combine MAE and MSE to make the reconstruction branch network pay more attention to the edge details of the region of interest between the spatially related images. $Loss_{B}$ is the loss function of the basic segmentation sub-network, and $Loss_{C}$ is the loss function of the complete segmentation sub-network. They are the same and are defined as follows:
\begin{equation}
Loss_{B} = Loss_{C} = weights * Loss_{CE} + Loss_{dice},
\label{eq:n10}
\end{equation}
\begin{equation}
Loss_{CE} = - \frac{1}{n} \sum_{i=1}^{n} y_{i} \ln{\hat{y_{i}}},
\label{eq:n11}
\end{equation}
\begin{equation}
Loss_{dice} = n - \sum_{i=1}^{n} \frac{2 \vert y_{i} \cap \hat{y_{i}} \vert}{\vert y_{i} \vert \cup \vert \hat{y_{i}} \vert},
\label{eq:n12}
\end{equation}
where $Loss_{CE}$ is the cross-entropy loss, and $Loss_{dice}$ is the dice loss proposed in \cite{68}. $weights$ is the loss weight for cross-entropy, which is set to $0.4$ in our method. Using cross-entropy loss and dice loss at the same time is to make the network focus more on segmenting the foreground area while reducing the difficulty of network training.

\section{Experiments and Results}

\subsection{Datasets}

We evaluate the performance of EAA-Net in 2020 Multimodal Brain Tumor Segmentation Challenge (BraTS 2020) \cite{81} and COVID-19 Lung CT Lesion Segmentation Challenge-2020 (COVID-19-20) \cite{82}. 

BraTS 2020 includes $369$ training samples and $125$ validation samples. All samples are 3D images, and each sample includes $155$ images with the size of $240\times240$. BraTS 2020 does not provide annotation for validation dataset. To ensure the fairness of the evaluation of brain tumor segmentation results, participants need to upload the segmentation results of the validation data to the online evaluation platform (CBICA's IPP) to obtain the evaluation results.

COVID-19-20 consists of $199$ training samples and $50$ validation samples. Similar to BraTS 2020, all samples are 3D images, and validation dataset does not contain annotation. The number of slices included in the sample is not fixed. The size of each slice is $512\times512$.

\begin{figure}[htbp]
	\centering
	\includegraphics[scale=0.15]{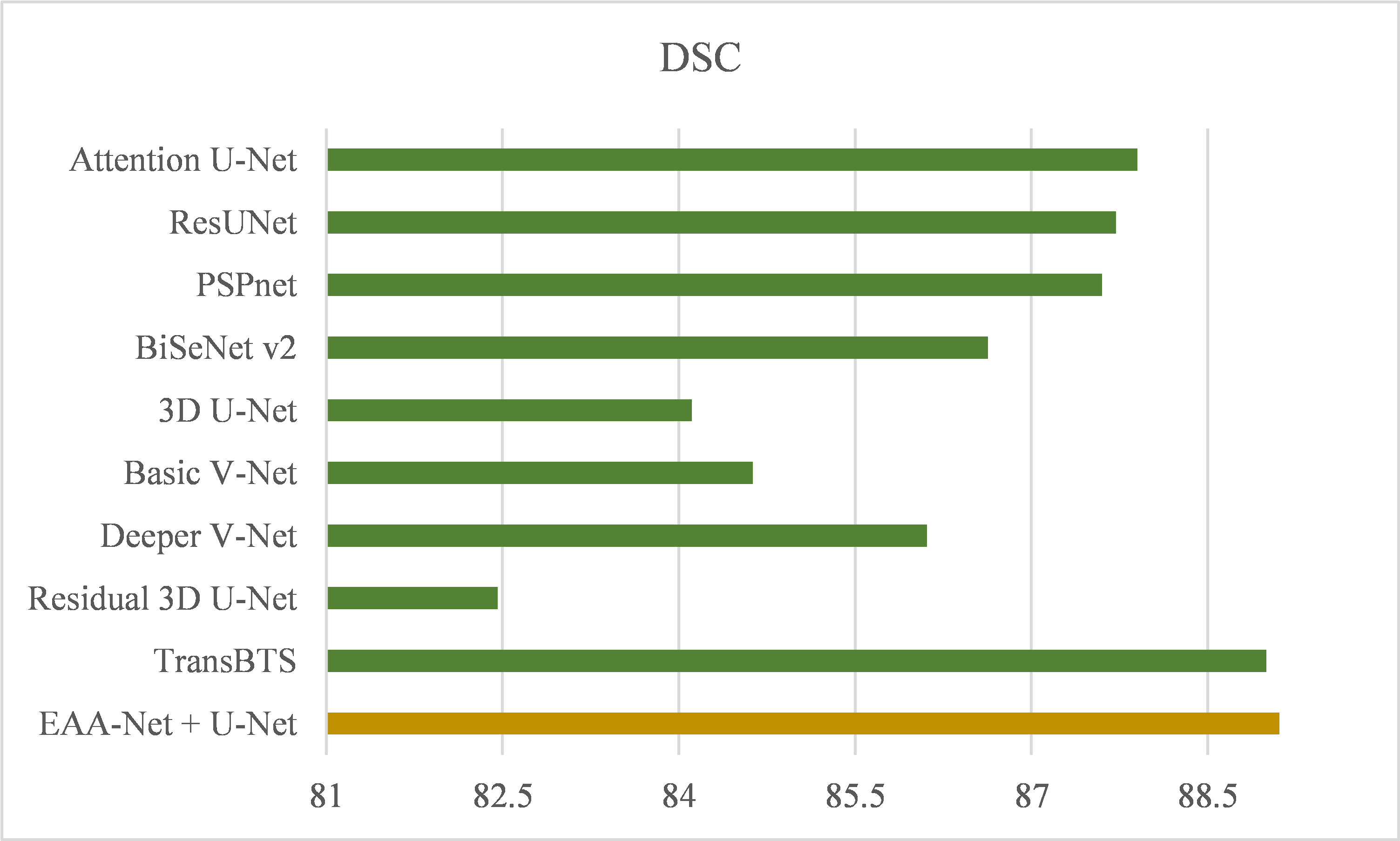}
	\caption{Comparison with the State of the Art on DSC of BraTS 2020.}
	\label{fig4}
\end{figure}

\subsection{Evaluation Metrics}

We evaluate the performance of the 3D image segmentation method EAA-Net by measuring the segmentation accuracy on a variety of datasets. Commonly used accuracy metrics for image segmentation include hausdorff distance (HD), Sensitivity, Specificity, dice similarity coefficient (DSC) and VolumeSimilarity (VS). The metrics are computed as follows:

\begin{equation}
\text{DSC} = \frac{2TP}{FP + 2TP + FN},
\label{eq:n13}
\end{equation}
\begin{equation}
\text{Sensitivity} = \frac{TP}{TP+FN},
\label{eq:n14}
\end{equation}
\begin{equation}
\text{Specificity} = \frac{TN}{TN+FP},
\label{eq:n15}
\end{equation}
\begin{equation}
\text{VolumeSimilarity} = \frac{2 * \lvert M \rvert - \lvert W \rvert}{\lvert M \rvert + \lvert W \rvert},
\label{eq:n16}
\end{equation}
\begin{equation}
\text{HD} = \max \{ \max_{u \in U} \min_{v \in V} (u,v), \max_{v \in V} \min_{u \in U} (u,v) \},
\label{eq:n17}
\end{equation}
where $TP$ is the number of true positives at the pixel-level, which is used to evaluate the performance for image segmentation methods. Similarly, $FP$, $TN$, $FN$ refer to false positives, true negatives, and false negatives respectively. HD calculates the similarity between the predicted ROI pixel distribution and the ground truth distribution. $u$ and $v$ respectively represent a point in the ground-truth set $U$ and the prediction set $V$. $M$ represents the set of voxels bounded by the manually delineated, and $W$ refer to the set of voxels bounded by the propagated contours.

\begin{table}[htpb]
\centering
\caption{Ablation study on the EAA-Net equipped with different multi-feature fusion modules on BraTS 2020. We show the DSC (\%), HD95, Sensitivity (\%) and Specificity (\%) on whole brain tumor.}
\resizebox{1.0\columnwidth}{!}{
    \begin{tabular}{lcccc}    
        \hline
        Module         & DSC          & HD95          & Sensitivity & Specificity \\
        \hline
        \hline        
        FFM \cite{58}  & 86.79        & 9.44          & 86.76       & 99.78 \\
        Element-wise Addition       & 88.38        & 5.76          & 87.34       & 99.90 \\ 
        Concatenation  & 88.52        & 6.18          & 87.68       & 99.90 \\
        AG \cite{18}   & 88.64        & 6.16          & 87.33       & {\bf 99.91} \\
        $I^{2}$ fusion module      & {\bf 89.11}  & {\bf 4.75}    & {\bf 88.44} & 99.90 \\
        \hline
    \end{tabular}
}
\label{table1}
\end{table}

\subsection{Implementation Details}

All experiments are implemented using TensorFlow 2.1.0. We train EAA-Net on an NVIDIA Tesla V100, the network uses Adam optimizer for training with 50 epochs. We use a batch size of $32$ to fill our GPU memory. The initial learning rate is $1e^{-4}$ and decrease it according to:
\begin{equation}
\alpha = \alpha_{0} * (1 - \frac{i}{N})^{0.9},
\label{eq:n18}
\end{equation}
where $\alpha_{0}$ is the initial learning rate, $N$ is the number of total epochs.

\subsection{Ablation Study on Key Components}

\begin{table}[htpb]
\centering
\caption{Performance on different Basic Segmentation networks on BraTS 2020. We show the DSC (\%), HD95 and parameters (M) of each model in the whole brain tumor segmentation.}
\resizebox{1.0\columnwidth}{!}{
    \begin{tabular}{lccc}    
        \hline
        Method                & DSC          & HD95          & Parameters\\
        \hline
        \hline
        U-Net                 & 86.82        & 8.86          & 7.38 \\ 
        EAA-Net + U-Net       & {\bf 89.11}  & {\bf 4.75}    & {\bf 2.38} \\
        \hline
        SegNet                & 87.30        & 7.22          & 8.46 \\
        EAA-Net + SegNet      & {\bf 88.39}  & {\bf 6.26}    & {\bf 2.90} \\
        \hline
        DeepLabv3+            & 88.37        & 6.95          & 11.90 \\
        EAA-Net + DeepLabv3+  & {\bf 88.64}  & {\bf 5.60}    & {\bf 11.09} \\ 
        \hline
    \end{tabular}
}
\label{table2}
\end{table}

\begin{table}[htpb]
\centering
\caption{Performance on different Basic Segmentation networks on COVID-19-20. We show the DSC (\%), VS (\%) and parameters (M) of each model in the lung lesion segmentation.}
\resizebox{1.0\columnwidth}{!}{
    \begin{tabular}{lccc}
       \hline
        Method                & DSC          & VS          & Parameters\\
        \hline
        \hline
        U-Net                 & 67.49        & 15.21        & 16.96 \\ 
        EAA-Net + U-Net       & {\bf 69.42}  & {\bf 24.16}  & {\bf 15.67} \\
        \hline
        SegNet                & 66.19        & 13.76        & 12.15 \\
        EAA-Net + SegNet      & {\bf 68.67}  & {\bf 16.85}  & {\bf 11.71} \\
        \hline
        DeepLabv3+            & 66.26        & 14.52        & 21.02 \\ 
        EAA-Net + DeepLabv3+  & {\bf 67.58}  & {\bf 15.28}  & {\bf 19.66} \\ 
        \hline
    \end{tabular}
}
\label{table3}
\end{table}

The $I^{2}$ fusion module is the key component of EAA-Net. To investigate the effectiveness of the $I^{2}$ fusion module, we perform ablation studies with other multi-feature fusion modules on BraTS 2020. Element-wise addition and concatenation are commonly used feature fusion modules in convolutional neural networks (CNN). Attention U-Net uses Attention Gate (AG) to integrate the low-level and high-level features, thereby alleviating the semantic gap between different levels of semantic information. Feature Fusion Module (FFM) is used to fuse all low-level and high-level features.

\begin{figure}[htbp]
	\centering
	\includegraphics[scale=0.15]{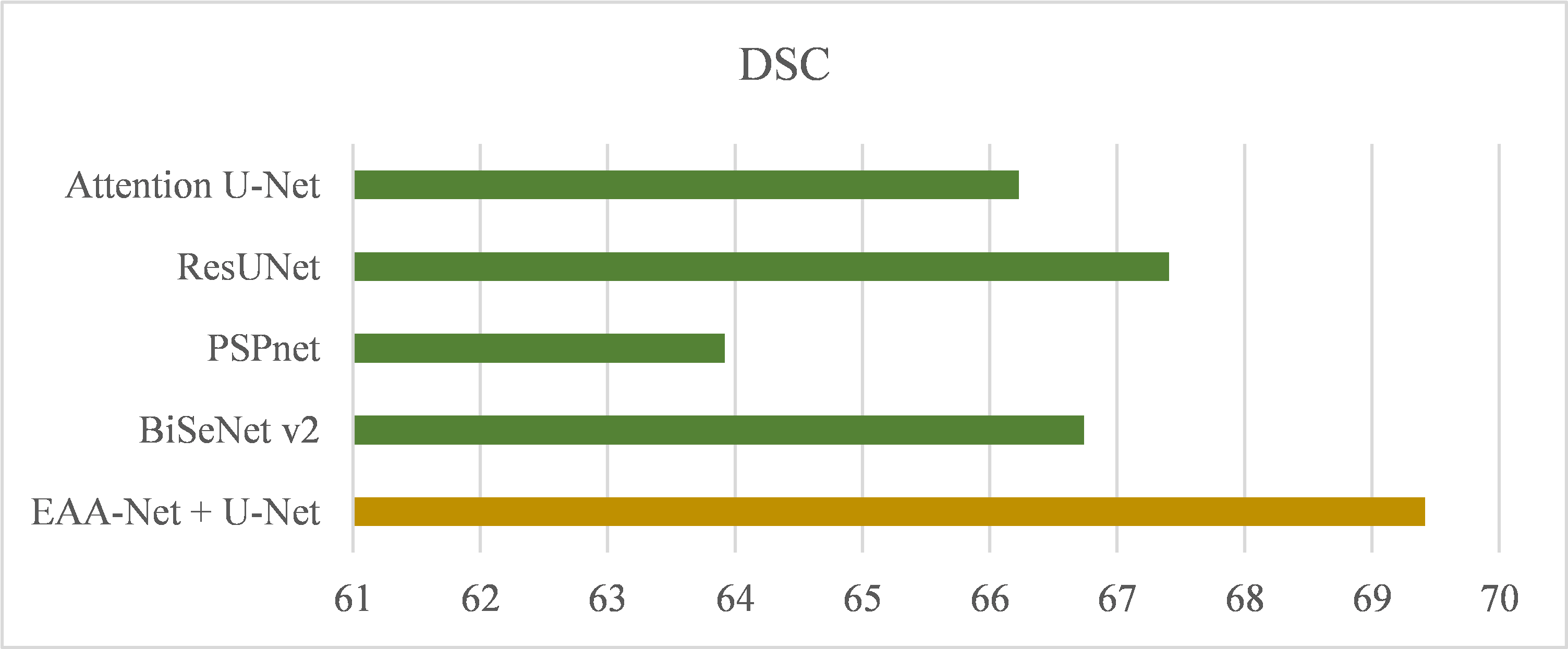}
	\caption{Comparison with the State of the Art on DSC of COVID-19-20.}
	\label{fig5}
\end{figure}

Comparison results are listed in Table \ref{table1}. The proposed $I^{2}$ fusion module achieved the best results on DSC, HD95 and Sensitivity. As shown in Table \ref{table1}, $I^{2}$ fusion module alleviate the difference in features between inter-class and intra-class. The $I^{2}$ fusion module greatly surpasses other multi-feature fusion modules on HD95, which indicates that the soft attention mechanism of the $I^{2}$ fusion module effectively reduces the influence of outliers caused by the intra-class features of the background region contained in the reconstruction network on the segmentation tasks.

\begin{table}[htpb]
\centering
\caption{Comparison with the State of the Art on BraTs 2020. We show the DSC (\%), HD95 and parameters (M) of each model in the whole brain tumor segmentation.}
\resizebox{1.0\columnwidth}{!}{
    \begin{tabular}{lccc}
        \hline
        Method               & DSC          & HD95        & Parameters\\
        \hline
        \hline
        Attention U-Net      & 87.90        & 7.37        & 9.25 \\ 
        ResUNet              & 87.72        & 7.44        & 2.78 \\
        PSPnet               & 87.60        & 5.95        & 9.33 \\
        BiSeNet v2           & 86.63        & 8.72        & 3.26 \\
        \hline
        3D U-Net \cite{80}            & 84.11        & 13.366      & 16.21 \\
        Basic V-Net \cite{80}         & 84.63        & 20.407      & - \\
        Deeper V-Net \cite{80}        & 86.11        & 14.499      & - \\
        Residual 3D U-Net \cite{80}   & 82.46        & 12.337      & - \\
        TransBTS \cite{80}   & 89.00        & 6.469       & 32.99 \\ 
        \hline
        EAA-Net + U-Net              & {\bf 89.11}  & {\bf 4.75}  & {\bf 2.38} \\ 
        \hline
    \end{tabular}
}
\label{table4}
\end{table}

\subsection{Performance on various Basic Segmentation Sub-networks}

We utilize U-Net \cite{4}, SegNet \cite{56} and DeepLabv3+ \cite{16} as the basic segmentation network of EAA-Net. These basic segmentation networks are variants of the autoencoder architecture. In this section, we compare the segmentation performance of these autoencoder architecture variants with EAA-Net that uses these variants as the basic segmentation network. We evaluate the general applicability of EAA-Net in improving the segmentation performance of the autoencoder architecture on BraTS 2020 and COVID-19-20.

Table \ref{table2} gives the segmentation results on BraTS 2020. Note that the segmentation performance of EAA-Net is better than corresponding autoencoder architecture. In the meantime, the number of parameters of EAA-Net is less than the corresponding autoencoder architecture. Table \ref{table2} shows that our method is superior to the segmentation methods without reconstruction sub-network on the HD95. Hausdorff distance is more sensitive to boundaries. Therefore, Hausdorff distance is often used for image segmentation tasks that pay attention to edge details. The Dice coefficient is the most frequently used metric in medical image competitions. It is a collective similarity metric, usually used to calculate the similarity of two samples. The Dice coefficient is more sensitive to the internal filling of the mask. The accurate segmentation of edge details also makes the dice coefficient of our framework higher than the corresponding backbone network. Experimental results indicate that EAA-Net significantly improves the segmentation efficiency of the autoencoder architectures. This trend also appears in lung lesion segmentation tasks, as shown in Table \ref{table3}. The qualitative comparison of the proposed EAA-Net is shown in Fig. \ref{fig2}. The edge segmentation results obtained by using EAA-Net are significantly better than the corresponding backbone network. This result not only appeared in the segmentation task of lung lesions, but also appeared in the segmentation task of brain tumors.

\begin{table}[htpb]
\centering
\caption{Comparison with the State of the Art on COVID-19-20. We show the DSC (\%), VS (\%) and parameters (M) of each model in the lung lesion segmentation.}
\resizebox{1.0\columnwidth}{!}{
    \begin{tabular}{lccc}
        \hline
        Method               & DSC          & VS           & Parameters\\
        \hline
        \hline
        Attention U-Net      & 66.23        & 13.34        & 17.24 \\ 
        ResUNet              & 67.41        & 14.08        & {\bf 15.50} \\
        PSPnet               & 63.92        & 1.04         & 20.94 \\
        BiSeNet v2           & 66.74        & -5.51        & 19.92 \\
        EAA-Net + U-Net      & {\bf 69.42 }  & {\bf 24.16}  & 15.67 \\ 
        \hline
    \end{tabular}
}
\label{table5}
\end{table}
 
\subsection{Comparison with the State of the Art}

In order to show the segmentation performance of EAA-Net intuitively, we compared our framework with popular semantic segmentation methods including Attention U-Net \cite{18}, ResUNet \cite{74}, PSPnet \cite{72} and BiSeNetv2 \cite{73}. To ensure fairness, we reproduce the state-of-the-art methods on the same training benchmark, and all results are provided by the online evaluation system. Furthermore, we compare our method with state-of-the-art medical image segmentation methods on BraTS 2020. These state-of-the-art medical image segmentation methods include 3D U-Net, Basic V-Net, Deeper V-Net, Residual 3D U-Net, and transformer-based TransBTS.

\begin{figure*}[htbp]
	\centering
	\includegraphics[scale=0.5]{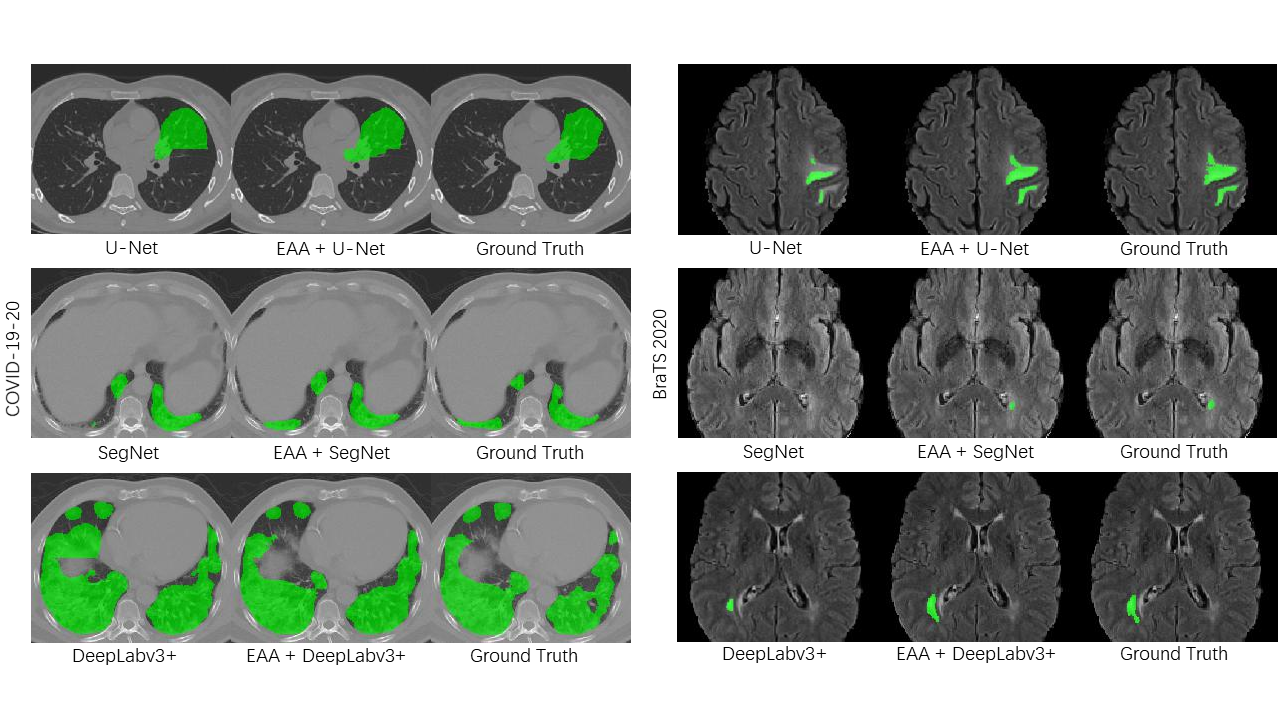}
	\caption{Qualitative comparison against other methods. The segmentation results of the lesion area are indicated in green color. Our method improve the quality of segmentation edges.}
	\label{fig3}
\end{figure*}

In Table \ref{table4} and Table \ref{table5}, our framework surpasses the state-of-the-art methods. It is worth noting that the number of parameters of our method is much smaller than other methods. 3D U-Net, Basic V-Net, Deeper V-Net, Residual 3D U-Net, Attention U-Net and ResUNet are commonly used medical image segmentation architectures. The improvement based on U-Net enables them to fuse edge information with high-level semantic information to obtain more accurate segmentation results. DSC, Hausdorff distance and VS are all sensitive to segmenting foreground details. The comparison results of our method with the above medical image segmentation methods demonstrate that our method has better performance in segmenting the details of the lesion area. However, BiSeNetv2 does not have an effective method to transmit low-level edge information to the prediction part, so its performance in medical image segmentation tasks is poor. The feature extraction method of PSPnet will lose a lot of edge details, which is unfavorable for medical image segmentation tasks that focus on edges. Furthermore, on edge details segmentation, the comparison with TransBTS proves that our method is still highly competitive with the transformer architecture which is without downsampling operation. Moreover, the number of parameters of EAA-Net is less than $1/13$ of TransBTS. Fig. \ref{fig4} and Fig. \ref{fig5} show the more intuitive comparison results on DSC.

\section{Conclusion}

In this paper, we propose EAA-Net, which is a light-weight autoencoder architecture for extracting edge information. EAA-Net extracts intra-class features through the reconstruction network, making the model focus on the edge details of the foreground in neighboring slices. EAA-Net is easy to implement and has small calculation cost. Depending on $I^{2}$ fusion module, EAA-Net merges the features of intra-class and inter-class, and reduces the influence of invalid intra-class features on segmentation tasks. The experimental results prove that our method can effectively improve the performance of the autoencoder architecture on medical image segmentation tasks. We hope this simple method will motivate people to rethink the roles of intra-class features for medical image segmentation.

\section{ACKNOWLEDGMENTS}

This work is supported by a grant from National Key R\&D Program of China (2020YFA0908400) and National Natural Science Foundation of China (NSFC 62172296, 61972280).

\bibliographystyle{IEEEbib}
\bibliography{IEEEbib}

\end{document}